\documentclass[preprint,12pt]{elsarticle}

\usepackage[cmex10]{amsmath}
\usepackage{amssymb}
\usepackage{amsthm}
\usepackage{algorithmic}
\usepackage{algorithm}
 \usepackage{tabularx}
\usepackage[toc,page]{appendix}
\usepackage{longtable}

\journal{Applied Soft Computing}

\begin{document}

\begin{frontmatter}



\title{An Outlier Detection-based Tree Selection Approach to Extreme Pruning of Random Forests}

 \author[label1]{Khaled Fawagreh}
\author[label1]{Mohamed Medhat Gaber}
\author[label1]{Eyad Elyan}
 \address[label1]{School of Computing Science and Digital Media\\
Robert Gordon University\\ \{k.fawagreh, m.gaber1, e.elyan\} @rgu.ac.uk \\Riverside East, Garthdee Road, \\ Aberdeen, AB10 7GJ, UK}


\begin{abstract}

Random Forest (RF) is an ensemble classification technique that was developed by Breiman over a decade ago. Compared with other ensemble techniques, it has proved its accuracy and superiority. Many researchers, however, believe that there is still room for enhancing and improving its performance in terms of predictive accuracy. This explains why, over the past decade, there have been many extensions of RF where each extension employed a variety of techniques and strategies to improve certain aspect(s) of RF. Since it has been proven empirically that ensembles tend to yield better results when there is a significant diversity among the constituent models, the objective of this paper is twofolds. First, it investigates how an unsupervised learning technique, namely, Local Outlier Factor (LOF) can be used to identify diverse trees in the RF.  Second, trees with the highest LOF scores are then used to produce an extension of RF termed \emph{LOFB-DRF} that is much smaller in size than RF, and yet performs at least as good as RF, but mostly exhibits higher performance in terms of accuracy. The latter refers to a known technique called ensemble pruning. Experimental results on 10 real datasets prove the superiority of our proposed extension over the traditional RF. Unprecedented pruning levels reaching as high as 99\% have been achieved at the time of boosting the predictive accuracy of the ensemble. The notably high pruning level makes the technique a good candidate for real-time applications.    

\end{abstract}


\end{frontmatter}


\section{Introduction}
\label{intro}
Ensemble classification is an application of ensemble learning to boost the accuracy of classification. Ensemble learning is a supervised machine learning paradigm where multiple models are used to solve the same problem \cite{polikar2006ensemble} \cite{rokach2010ensemble} \cite{kuncheva2003measures}. Since single classifier systems have limited predictive performance \cite{yan2004designing} \cite{polikar2006ensemble} \cite{maclin2011popular}  \cite{rokach2010ensemble}, ensemble classification was developed to yield better predictive performance \cite{polikar2006ensemble} \cite{maclin2011popular}  \cite{rokach2010ensemble}. In such an ensemble, multiple classifiers are used. In its basic mechanism, majority voting is then used to determine the class label for unlabeled instances where each classifier in the ensemble is asked to predict the class label of the instance being considered. Once all the classifiers have been queried, the class that receives the greatest number of votes is returned as the final decision of the ensemble.

Three widely used ensemble approaches could be identified, namely, boosting, bagging, and stacking. Boosting is an incremental process of building a sequence of classifiers, where each classifier works on the incorrectly classified instances of the previous one in the sequence. AdaBoost \cite{freund1997decision} is the representative of this class of techniques. However, AdaBoost is proned to overfitting. The other class of ensemble approaches is the Bootstrap Aggregating (Bagging) \cite{breiman1996baggingrandom}. Bagging involves building each classifier in the ensemble using a randomly drawn sample of the data with replacement, having each classifier give an equal vote when labeling unlabeled instances. Bagging is known to be more robust than boosting against model overfitting. Random Forest (RF) is the main representative of bagging  \cite{breiman2001random}. Stacking (sometimes called stacked generalization) extends the cross-validation technique that partitions the data set into a held-in data set and a held-out data set; training the models on the held-in data; and then choosing whichever of those trained models performs best on the held-out data. Instead of choosing among the models, stacking combines them, thereby typically getting performance better than any single one of the trained models \cite{wolpert1992stacked}. Stacking has been successfully used in both supervised learning tasks (regression) \cite{breiman1996stacked}, and unsupervised learning (density estimation) \cite{smyth1999linearly}. 

The ensemble method that is relevant to our work in this paper is RF. RF has been proved to be the state-of-the-art ensemble classification technique. Since RF algorithms typically build between 100 and 500 trees \cite{williams2011use}, it would be useful to reduce the number of trees participating in majority voting and yet achieving better performance both in terms of accuracy and speed. In this paper, we propose an unsupervised learning approach to improve speed and accuracy of RF. For speed, our approach avoids having all trees participate in majority voting as only a small subset of the trees is selected. For accuracy, since it has been proven empirically that ensembles tend to yield better results when there is a significant diversity among the models \cite{kuncheva2003measures}  \cite{brown2005diversity} \cite{adeva2005accuracy} \cite{tang2006analysis}, our approach ensures that diverse trees in the ensemble are selected. We adopted Local Outlier Factor for tree diversification. Hence, the method is termed Local Outlier Factor Based Diversified Random Forest (both \emph{LOFB-DRF} and \emph{LOF-DRF} are used interchangeably) . 

This paper is organized as follows. First we discuss related work in Section \ref{related}. This is followed by Section \ref{prem} where the motivation and an introduction to RF are covered.  Section \ref{lof} describes the Local Outlier Factor that will be utilized in our proposed extension of RF. Section \ref{LOFB-DRF} formalizes our proposed method and corresponding algorithm. Experimental study demonstrating the superiority of the proposed technique over the traditional RF is detailed in Section \ref{experiment}. The paper is then concluded with a summary and pointers to future directions in Section \ref{conc}.  

\section{Related Work}
\label{related}

Several attempts have been made in recent years in order to produce a subset of an ensemble  that performs as well as, or better than, the original ensemble. The purpose of ensemble pruning is to search for such a good subset. This is particularly useful for large  ensembles that require extra memory usage, computational costs, and occasional decreases in effectiveness.  Grigorios et al. \cite{tsoumakas2009ensemble} recently amalgamated a survey of ensemble pruning techniques where they classified such techniques into four categories: ranking based, clustering based, optimization based, and others. Ranking based methods, that are relevant to us in this paper, are conceptually the simplest. Since using the predictive performance to rank models is too simplistic and does not yield satisfying results \cite{partridge1996engineering} \cite{yang2005ensemble}, ranking based methods employ an evaluation measure to rank models. Kappa statistic measure $\kappa$ was used in \cite{margineantu1997pruning} for pruning AdaBoost ensembles.  For  bagging ensembles, however, kappa has proven to be non-competitive \cite{martinez2009analysis}. For bagging ensembles, \cite{martinez2006pruning} developed an efficient and effective pruning method based on orientation ordering where the classifiers obtained from bagging are reordered  and a subset is selected for aggregation. 

An interesting issue that remains after ranking the models is to determine the models that will be chosen to form the pruned ensemble.  For this,  two approaches can be used. The first approach is to use a fixed user-specified amount or percentage of models. A second approach is to dynamically select the size based on the evaluation measure or the predictive performance of ensembles of different sizes.
In this paper, the models will be ranked according to their Local Outlier Factor (LOF) values and the models with the top k (where k is a multiple of 5 ranging from 5 to 40) values will be selected to form the pruned  ensemble.

\subsection{Diversity Creation Methods}
Because of the vital role diversity plays on the performance of ensembles, it had received a lot of attention from the research community. G. Brown et al. \cite{brown2005diversity} summarized the work done to date in this domain from two main perspectives. The first is a review of the various attempts that were made to provide a formal foundation of diversity. The second, which is more relevant to this paper, is a survey of the various techniques to produce diverse ensembles. For the latter, two types of diversity methods were identified: implicit and explicit. While implicit methods tend to use randomness to generate diverse trajectories in the hypothesis space, explicit methods, on the other hand, choose different paths in the space deterministically. In light of these definitions, bagging and boosting in the previous section are classified as implicit and explicit respectively. 

G. Brown et al. \cite{brown2005diversity} also categorized ensemble diversity techniques into three categories: starting point in hypothesis space, set of accessible hypotheses, and manipulation of training data. Methods in the first category use different starting points in the hypothesis space, therefore, influencing the convergence place within the space.  Because of their poor performance of achieving diversity, such methods are used by many authors as a default benchmark for their own methods \cite{maclin2011popular}. Methods in the second category vary the set of hypotheses that are available and accessible by the ensemble. For different ensembles, these methods vary either the training data used or the architecture employed. In the third category, the methods alter the way space is traversed. Occupying any point in the search space, gives a particular hypothesis. The type of the ensemble obtained will be determined by how the space of the possible hypotheses is traversed.

In this paper, we propose a new diversity creation method based on unsupervised learning. The method utilizes an existing unsupervised learning technique that, to the best of our knowledge, has not been used before in the production of pruned ensembles.

\subsection{Diversity Measures}
Regardless of the diversity creation technique used, diversity measures were developed to measure the diversity of a certain technique or perhaps to compare the diversity of two techniques. Tang et al. \cite{tang2006analysis} presented a theoretical analysis on six existing diversity measures: disagreement measure  \cite{skalak1996sources}, double fault measure  \cite{giacinto2001design}, KW variance \cite{kohavi1996bias}, inter-rater agreement \cite{fleiss2013statistical}, generalized diversity \cite{partridge1997software}, and measure of difficulty \cite{fleiss2013statistical}. The goal was not only to show the underlying relationships between them, but also to relate them to the concept of margin, which is one of the contributing factors to the success of ensemble learning algorithms.

We suffice to describe the first two measures as the others are outside the scope of this paper. The disagreement measure is used to measure the diversity between two base classifiers \emph{$h_j$} and \emph{$h_k$}, and is calculated as follows:

$$
dis_{j,k} = \frac{N^{10}+N^{01}}{N^{11}+N^{10}+N^{01}+N^{00}}
$$
\\
where \\
\begin{itemize}
\item \emph{$N^{10}$}: means number of training instances that were correctly classified by \emph{$h_j$}, but are incorrectly classified by \emph{$h_k$}
\item \emph{$N^{01}$}: means number of training instances that were incorrectly classified by \emph{$h_j$}, but are correctly classified by \emph{$h_k$} 
\item \emph{$N^{11}$}: means number of training instances that were correctly classified by  \emph{$h_j$} and  \emph{$h_k$}
\item \emph{$N^{00}$}: means number of training instances that were incorrectly classified by  \emph{$h_j$} and  \emph{$h_k$} 
\end{itemize}
The higher the disagreement measure, the more diverse the classifiers are. The double fault measure uses a slightly different approach where the diversity between two classifiers is calculated as:
$$
DF_{j,k} = \frac{N^{00}}{N^{11}+N^{10}+N^{01}+N^{00}}
$$

 The above two diversity measures work only for binary classification (AKA binomial) where there are only two possible values (like Yes/No) for the class label, hence, the objects are classified into exactly two groups.  They do not work for multiclass (AKA multinomial) classification where the objects are classified into more than two groups.

\section{Preliminaries}
\label{prem}
\subsection{Motivation}
\label{motive}
As mentioned before, RF algorithms tend to build between 100 and 500 trees \cite{williams2011use}. Our research aims at producing child RFs that are significantly smaller in size and yet, have accuracy performance that is at least as good as that of the parent RF from which they were derived. The classification speed of each child is guaranteed to be much faster than that of the parent RF because 1) it has much fewer trees and 2) any tree used in the child is also in the parent (i.e., no new trees were introduced in the child). 

\subsection{Random Forest}
\label{rf}
RF is an ensemble learning method used for classification and regression. Developed by Breiman \cite{breiman2001random}, the method combines Breiman's bagging sampling approach \cite{breiman1996baggingrandom}, and the random selection of features, introduced independently by Ho \cite{ho1995random} \cite{ho1998random} and Amit and Geman \cite{amit1997shape}, in order to construct a collection of decision trees with controlled variation. Using bagging,  each decision tree in the ensemble is constructed using a sample with replacement from the training data. Statistically, the sample is likely to have about 64\% of instances appearing at least once in the sample. Instances in the sample are referred to as in-bag-instances, and the remaining instances (about 36\%), are referred to as out-of-bag instances. Each tree in the ensemble acts as a base classifier to determine the class label of an unlabeled instance. This is done via majority voting where each classifier casts one vote for its predicted class label, then the class label with the most votes is used to classify the instance. Algorithm \ref{rfalgo} below depicts the RF algorithm \cite{breiman2001random} where N is the number of training samples and S is the number of features in data set. 

\begin{algorithm}[!htb]
\caption{Random Forest Algorithm}          
\label{rfalgo}                           
\begin{algorithmic}
          
\STATE \COMMENT{User Settings}
\STATE input $N$, $S$   
\STATE \COMMENT{Process}
\STATE Create an empty vector $\overrightarrow{RF}$ 
\FOR{$i = 1 \to N$}
\STATE Create an empty tree $T_i$
\REPEAT
\STATE Sample $S$  out of all features $F$ using Bootstrap sampling 
\STATE Create a vector of the $S$ features $\overrightarrow{F_S}$
\STATE Find Best Split Feature $B(\overrightarrow{F_S})$
\STATE Create A New Node using $B(\overrightarrow{F_S})$ in $T_i$
\UNTIL{No More Instances To Split On}
\STATE Add $T_i$ to the $\overrightarrow{RF}$ 
\ENDFOR
\STATE \COMMENT{Output}
\STATE A vector of trees $\overrightarrow{RF}$
\end{algorithmic}
\end{algorithm}


\section{Local Outlier Factor}
\label{lof}
The Local Outlier Factor (LOF) algorithm was developed by Breunig et al.  \cite{breunig2000lof} to measure the outlierness of an object. The higher the LOF value assigned to an object, the more  isolated the object is with respect to its neighbors. It  is considered a very powerful anomaly detection technique in machine learning and classification. Earlier work on outlier detection was investigated in \cite{arning1996linear} \cite{ruts1996computing} \cite{knox1998algorithms} \cite{knorr1999finding}, however, the work was limited by treating an outlier as a binary property to classify an object as an outlier or not, without assigning it a value to measure its outlierness as was done in  \cite{breunig2000lof}. 

The LOF can be used as a method to achieve diversity. It was one of 3 strategies used to obtain diversity when constructing an ensemble for the KDDCup 1999 dataset  \cite{erich2011interpreting}. Schubert et al. \cite{schubert2012evaluation} proposed methods for measuring similarity and diversity of methods for building advanced outlier detection ensembles using LOF variants and other algorithms.

Formally,  Breunig et al. \cite{breunig2000lof} introduced the concept of reachability distance in order to calculate the LOF. If the distance of object $A$ to the \emph{k} nearest neighbor is denoted by  k{-}distance(A), where the \emph{k} nearest neighbors is denoted by $N_{k}$(A), the following equation defines the reachability distance (rd):

\begin{equation}
rd_{k}(A,B) = max\{k{-}distance(B),d(A,B)\}
\end{equation}
where $d(A,B)$ is the distance between objects $A$ and $B$. The local reachability density of object $A$ is then defined by 

\begin{equation}
lrd(A) = \frac{\sum_{B \in N_{k}(A)} rd_{k}(A,B)}{|N_k(A)|}
\end{equation}
Using the local reachability density of object $A$ as defined in the previous equation, the  LOF for object $A$ is given by:

\begin{equation}
LOF_k(A) = \frac{\sum_{B \in N_{k}(A)} \frac{lrd(B)}{lrd(A)}}{|N_k(A)|}
\end{equation}
 
\section{LOF-Based Diverse Random Forest (LOFB-DRF)}
\label{LOFB-DRF}

\begin{figure}[t]
\centering
\includegraphics[width=\columnwidth]{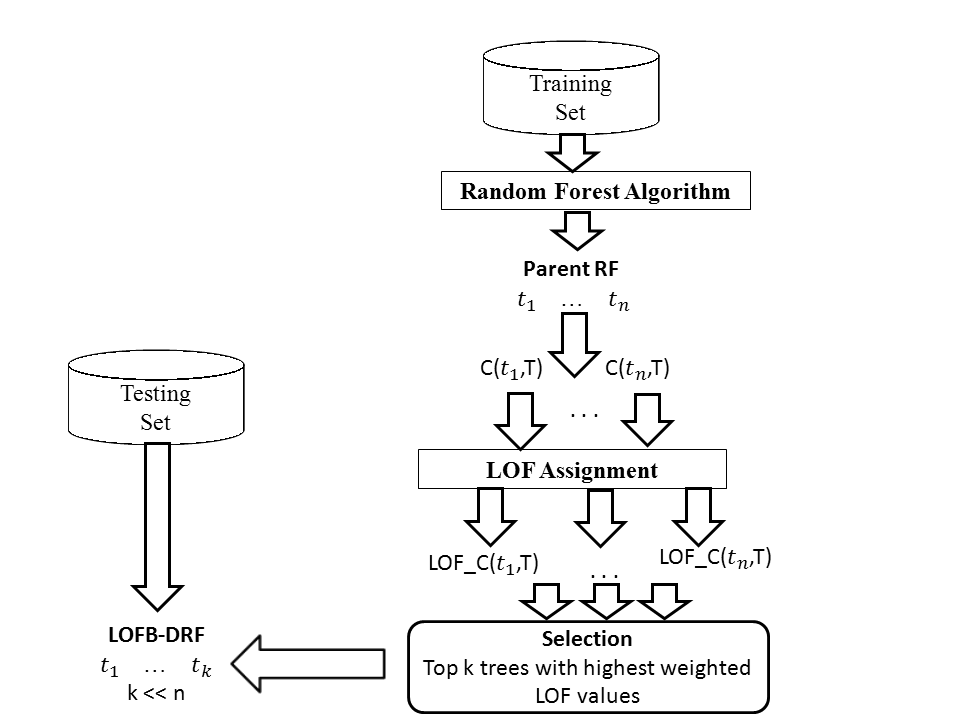}
\caption{LOFB-DRF Approach}
\label{lofdrf}
\end{figure}

In this section, we propose an extension of RF called LOFB-DRF that spawns a child RF that is 1) much smaller in size than the parent RF and 2) has an accuracy that is at least as good as that of the parent RF. In this extension, we use the LOF discussed in Section \ref{lof}.  As shown in Figure \ref{lofdrf}, each tree predictions on the training dataset (denoted by the vector $C(t_{i},T)$) is assigned an LOF value that indicates the degree of its  outlierness. The top k (k=5,10,...,40) trees corresponding to these predictions  with the highest weighted LOF values (to be discussed next) are then  selected to become members of the resulting LOFB-DRF.  In the remainder of this paper, we will refer to the parent/original traditional Random Forest as \emph{RF}, and refer to the resulting child RF based on our method as \emph{LOFB-DRF}.

Based on Figure \ref{lofdrf}, we formalize the LOFB-DRF algorithm as shown in Algorithm \ref{lof-drf-algo} where $T$ is the training set. The constant $k$  refers to the number of trees that will have the highest weighted LOF values as will be discussed later. The domain of this constant is multiple of 5  in the range 5 to 40. This way and as we shall see in the experiments section, we can compare the performance RF with an LOFB-DRF of different sizes.

It is important to remember that the size of the resulting LOFB-DRF is determined by the constant $k$. For example,  if $k$ is 5, then the resulting LOFB-DRF will have size 5, and so on. 

\begin{algorithm}[!htb]
\caption{LOFB-DRF Algorithm}          
\label{lof-drf-algo}                           
\begin{algorithmic}
          
\STATE \COMMENT{User Settings}
\STATE input $T$, $k$
\STATE \COMMENT{Process}
\STATE Create an empty vector $\overrightarrow{treesPredictions}$ 
\STATE Create an empty vector $\overrightarrow{LOFB-RF}$ 
\STATE Using \emph{T}, call Algorithm $\ref{rfalgo}$ above to create the parent  \emph{RF} 
\FOR{$i = 1 \to RF.getNumTrees()$}
\STATE  $\overrightarrow{treesPredictions}$ $\Leftarrow$  $\overrightarrow{treesPredictions}$ $\cup$ C(RF.tree(i), T)
\ENDFOR
\STATE For each instance in $\overrightarrow{treesPredictions}$, assign an LOF value
\STATE Select the top $k$  instances in $\overrightarrow{treesPredictions}$ with highest weighted LOF values
\STATE Select the corresponding  trees from RF and add them to $\overrightarrow{LOFB-DRF}$
\STATE \COMMENT{Output}
\STATE A vector of trees $\overrightarrow{LOFB-DRF}$
\end{algorithmic}
\end{algorithm}

\subsection{Selection of Trees}
\label{wlof}
With reference to Algorithm \ref{lof-drf-algo}, the selection of trees in RF that will become members of LOFB-DRF proceeds as follows. First, predictions of each tree on the training dataset T is computed as a vector and added to the vector $\overrightarrow{treesPredictions}$. At the conclusion of the \textbf{for} loop, $\overrightarrow{treesPredictions}$ becomes a super vector containing vectors where each vector stores the predictions of each tree. Each instance in  $\overrightarrow{treesPredictions}$ is then assigned a normalized  LOF value  between 0 and 1. This way, each normalized value describes the probability of the instance being an outlier \cite{erich2011interpreting}.  Then we assign to each instance a weight that is the product of the normalized LOF value and the accuracy rate of the corresponding tree on the training data. Formally, let $c_i$ be an instance in the super vector $\overrightarrow{treesPredictions}$,  NormalizedLOF($c_i$) be the normalized LOF value assigned to this instance, and  AccuracyRate(Tree($c_i$),T) be the accuracy rate of Tree($c_i$) on the training dataset T where Tree($c_i$) is the tree that corresponds to the instance $c_i$. The weight assigned to this instance is given by:

\begin{equation}
weight = NormalizedLOF(c_i) \times AccuracyRate(Tree(c_i),T)
\end{equation}
The instances are then sorted in descending order by this weight and the corresponding top $k$ trees are then selected.

\subsection{Diversity Measure}

Here we propose a simple diversity measure to measure the diversity of classifiers that works with binary and multiclass classification. Given two classifiers \emph{$h_j$} and \emph{$h_k$} and a training set T of size \emph{n}. Let $C$($t_l$,$s_i$)  denotes the class label obtained  after having  $t_l$ classify the sample $s_i$ in the training set T. The diversity between the two classifiers can be measured by: 
\begin{equation}
\label{diveq}
diversity_{j,k}=\frac{\sum\limits^{n}_{i=1}\delta(C( t_j,c_i) ,C( t_k,c_i) )}{n} 
\end{equation}

where

\begin{equation}
    \delta(x_{j},y_{j})=
    \begin{cases}
      0, & \text{if}\ x_{j} = y_{j} \\
      1, & \text{otherwise}
    \end{cases}
  \end{equation}

The higher the number of discrepancies between the two classifiers, the higher the diversity is. For example, assume that we have a training set consisting of 10 training samples
T=\{$s_1$,$ s_2$,$ s_3$,$ s_4$,$ s_5$,$ s_6$,$ s_7$,$ s_8$,$ s_9$,$s_{10}$\}, and two classifiers $t_1$ and $t_2$. Assume also that there are 3 possible values for the class label \{a,b,c\}.  Let  C($t_1$,T)=\textless a,a,b,c,c,a,b,c,b,b\textgreater  
\hphantom aand C($t_2$,T)=\textless a,a,b,b,a,a,b,c,c,c\textgreater. According to \ref{diveq} above, the diversity between the two classifiers is therefore 4/10 or 40\%.

\section{Experiments}
\label{experiment}

For our experiments, we have used 10 real datasets with varying characteristics from the UCI repository \cite{Bache+Lichman:2013}. To use the holdout testing method, each dataset was divided into 2 sets: training and testing. Two thirds (66\%) were reserved for training and the rest (34\%) for testing.  Each dataset consists of input variables (features) and an output variable. The latter refers to the class label whose value will be predicted in each experiment.  For the RF in Figure \ref{lofdrf}, the initial RF to produce the LOFB-DRF had a size of 500 trees, a typical upper limit setting for RF \cite{williams2011use}. 

The LOFB-DRF algorithm described above was implemented using the Java programming language utilizing the API of Waikato Environment for Knowledge Analysis (WEKA) \cite{witten2005data}. We ran this algorithm 10 times on each dataset where a new RF was created in each run. We then calculated the average of the 10 runs for each resulting LOFB-DRF to produce the average for a variety of metrics including accuracy rate, minimum accuracy rate, maximum accuracy rate, standard deviation, FMeasure, and AUC as shown in Table \ref{weightedlofontraintable}.  For the RF, we just calculated the average accuracy rate, FMeasure, and AUC as shown in the last 3 columns of the table.

\subsection{Results} 
\label{result}

Table \ref{weightedlofontraintable} compares the performance of LOFB-DRF and RF on the 10 datasets used in the experiment. To show the superiority of LOFB-DRF, we have highlighted in boldface the average accuracy rate of LOFB-DRF when it is greater than that of RF.
With the exception of the \emph{audit} and \emph{vote} datasets (last 2 datasets), we find that LOFB-DRF performed at least as good as RF. Interestingly enough, of the 10 datasets, LOFB-DRF, regardless of its size, completely outperformed RF on 3 of the datasets, namely, \emph{squash-stored}, \emph{eucalyptus}, and \emph{sonar}. 
While LOFB-DRF lost to RF on only 2 datasets (\emph{audit} and  \emph{vote}), the difference was by a very small negligible fraction of less than 1\% (in the case of \emph{audit}), and less than 1.2\% (in the case of \emph{vote})!

\subsection{Pruning Level}

In ensemble pruning, a pruning level refers to the reduction ratio between the original ensemble and the pruned one. For example, if the size of the original ensemble is 500 trees and the pruned one is of size 50, then $100\% - \frac{50}{500} \times 100\%=90\%$ is the pruning level that was achieved in the pruned ensemble. This means that the pruned ensemble is 90\% smaller than the original one. Table \ref{PrunLevel} shows the pruning levels where the first column shows the maximum possible pruning level for an LOFB-DRF that has outperformed RF, and the second column shows the pruning level of the best performer LOFB-DRF. We can see that with extremely healthy pruning levels ranging from 95\% to 99\%, our technique outperformed RF. This makes LOFB-DRF a natural choice for real-time applications, where fast classification is an important desideratum. In most cases, 100 times faster classification can be achieved with the 99\% pruning level, as shown in the table. In the worst case scenario, only 16.67 times faster classification with 95\% pruning level in the \emph{squash-unstored} dataset. Such estimates are based on the fact that the number of trees traversed in the RF is the dominant factor in the classification response time. This is especially true, given that RF trees are unpruned bushy trees.   

Note that the \emph{audit} and \emph{vote} datasets were not listed in the table as the RFs for these datasets (refer to the last 2 datasets in Table \ref{weightedlofontraintable}) outperformed all LOFB-DRFs, however, by a very small amount as shown in Table \ref{range}.


\begin{table}[ht]
\caption{Maximum Pruning Level with Best Possible Performance} 
\label{PrunLevel}
\scalebox{0.9}{
\centering 
\begin{tabular}{l c c c} 
\hline\hline 
Dataset & Maximum Pruning Level & Best Performance Pruning Level \\ [0.5ex] 
\hline 
breast-cancer & 97\% & 95\%   \\ 
squash-unstored & 95\%  & 93\%   \\ 
squash-stored & 99\%  & 98\%   \\
eucalyptus & 99\%  & 99\%   \\
soybean & 98\%  & 97\%   \\
diabetes & 96\%  & 96\%   \\
car & 99\%  & 99\%   \\
sonar & 99\%  & 99\%   \\
\hline 
\end{tabular}}
\label{table:nonlin} 
\end{table}

\subsection{Analysis} 
\label{analysis}

\begin{figure}[H]
\centering
\scalebox{1.0}{

\includegraphics[width=6in,height=4in]{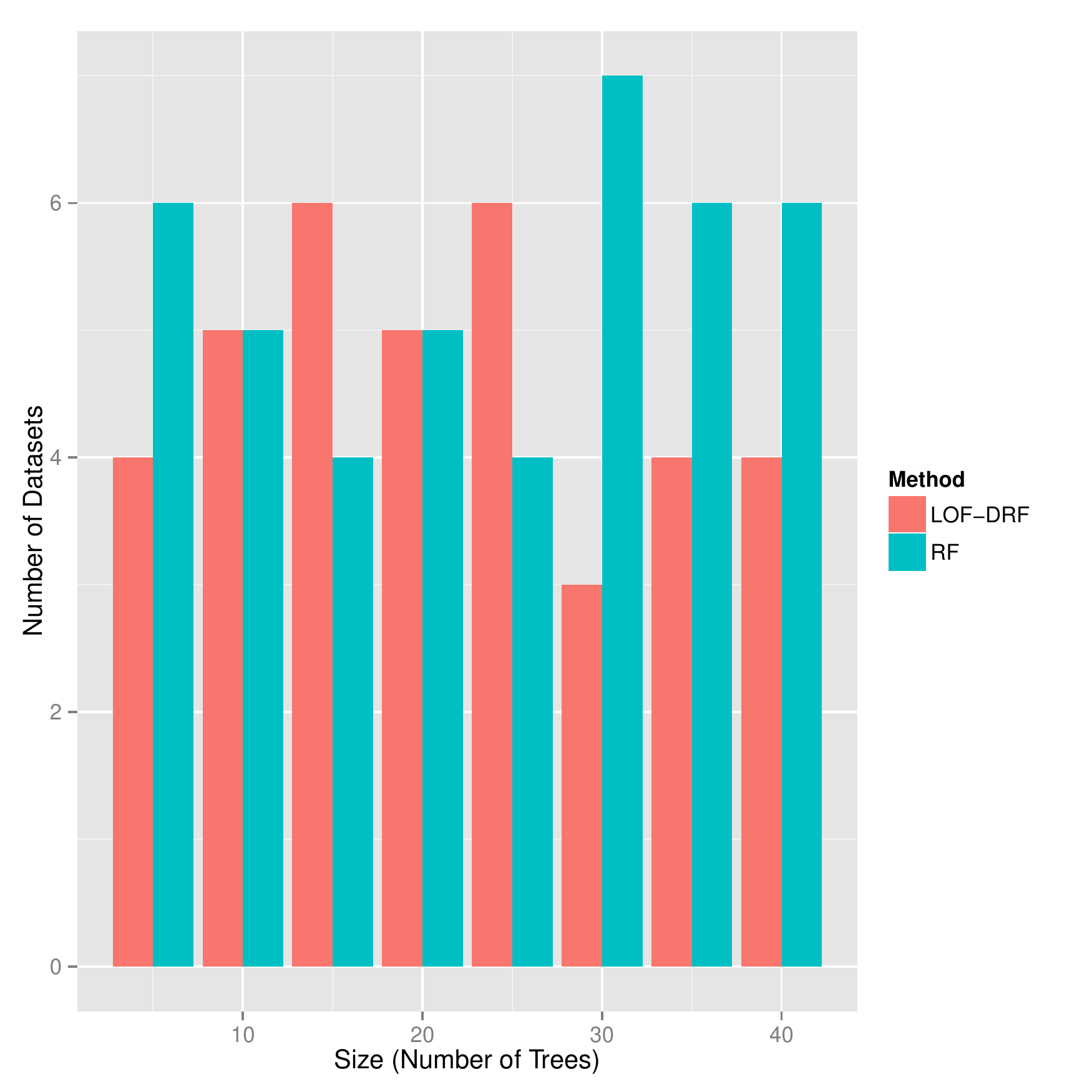}}
\caption{Accuracy Rate Comparison of RF \& LOFB-DRF}
\label{fig2}
\end{figure}

By showing the number of datasets each was superior on, Figure \ref{fig2} compares the accuracy rate of RF and LOFB-DRF using different sizes of LOFB-DRF. For sizes 10, 15, 20, and 25, the figure clearly shows that LOFB-DRF indeed performed at least as good as RF. As shown in Table \ref{range} below, for the cases (size 5, 30, 35, and 40) where RF outperformed LOFB-DRF, the difference was very small, considering the pruning level that was achieved. 

\begin{table}[lh]
\centering
\caption{Outperformance Range of RF Over LOFB-DRF}
\label{range}
\begin{tabular}{|c|c|c|c|c|c|}
\hline
 LOFB-DRF Size  & 5    &     30    &    35      & 40    \\
 \hline
Range & 0.31\% - 4.12\%  &	0.08\% - 2.78\% &	0.05\% - 1.45\% &	0.31\% - 3.33\% \\
 \hline
Pruning Level & 99\% & 94\% & 93\% & 92\% \\
\hline
\end{tabular}
\end{table}

\subsection{Bias/Variance Analysis}
\label{bandv}
Bias and variance are measures used to estimate the accuracy of a classifier \cite{kohavi1995study}.  The bias measures the difference between the classifier's predicted class value and the true value of the class label being predicted. The variance, on the other hand, measures the variability of the classifier's prediction as a result of sensitivity due to fluctuations in the training set. If the prediction is always the same regardless of the training set, it equals zero. However, as the prediction becomes more sensitive to the training set, the variance tends to increase. For a classifier to be accurate,  it should maintain a low bias and variance. 

There is a trade-off between a classifier's ability to minimize bias and variance. Understanding these two types of measures can help us diagnose classifier results and avoid the mistake of over- or under-fitting. Breiman et al. \cite{leo1984classification} provided an analysis of complexity and induction in terms of a trade-off between bias and variance. 
In this section, we will show that LOFB-DRF can have a bias and variance comparable to and even better than RF. Starting with bias, the first column in Table \ref{biasPrunLevel} shows the pruning level of LOFB-DRF that performed the best relative to RF, and the second column shows the pruning level of the smallest LOFB-DRF that outperformed RF. As demonstrated in the table, LOFB-DRF has outperformed RF on all datasets. On the other hand, Table \ref{variancePrunLevel} shows similar results but variance-wise. Once again, LOFB-DRF has outperformed RF on all datasets. Although looking at bias in isolation of variance (and vice versa) provides only half of the picture, our aim is to demonstrate that with a pruned ensemble, both bias and/or variance can be enhanced. We attribute this to the high diversity our ensemble exhibits. 

\begin{table}[ht]
\caption{Pruning Level for LOFB-DRF Bias} 
\label{biasPrunLevel}
\scalebox{0.9}{
\centering 
\begin{tabular}{l c c c} 
\hline\hline 
Dataset & Best Performer & Smallest LOFB-DRF Outperforming RF \\ [0.5ex] 
\hline 
breast-cancer & 99\% & 99\%   \\ 
squash-unstored & 94\%  & 95\%   \\ 
squash-stored & 98\%  & 99\%   \\
eucalyptus & 99\%  & 99\%   \\
soybean & 97\%  & 97\%   \\
diabetes & 99\%  & 99\%   \\
car & 94\%  & 97\%   \\
sonar & 92\%  & 99\%   \\
audit & 93\%  & 98\%   \\
vote & 98\%  & 99\%   \\ 
\hline 
\end{tabular}}
\label{table:nonlin} 
\end{table}

\begin{table}[ht]
\caption{Pruning Level for LOFB-DRF Variance} 
\label{variancePrunLevel}
\scalebox{0.9}{
\centering 
\begin{tabular}{l c c c} 
\hline\hline 
Dataset & Best Performer & Smallest LOFB-DRF Outperforming RF \\ [0.5ex] 
\hline 
breast-cancer & 95\% & 99\%   \\ 
squash-unstored & 99\%  & 99\%   \\ 
squash-stored & 97\%  & 97\%   \\
eucalyptus & 93\%  & 98\%   \\
soybean & 94\%  & 94\%   \\
diabetes & 98\%  & 98\%   \\
car & 99\%  & 99\%   \\
sonar & 99\%  & 99\%   \\
audit & 97\%  & 97\%   \\
vote & 92\%  & 99\%   \\ 
\hline 
\end{tabular}}
\label{table:nonlin} 
\end{table}

\begin{figure}[H]
\centering
\scalebox{1.0}{

\includegraphics[width=6in,height=4in]{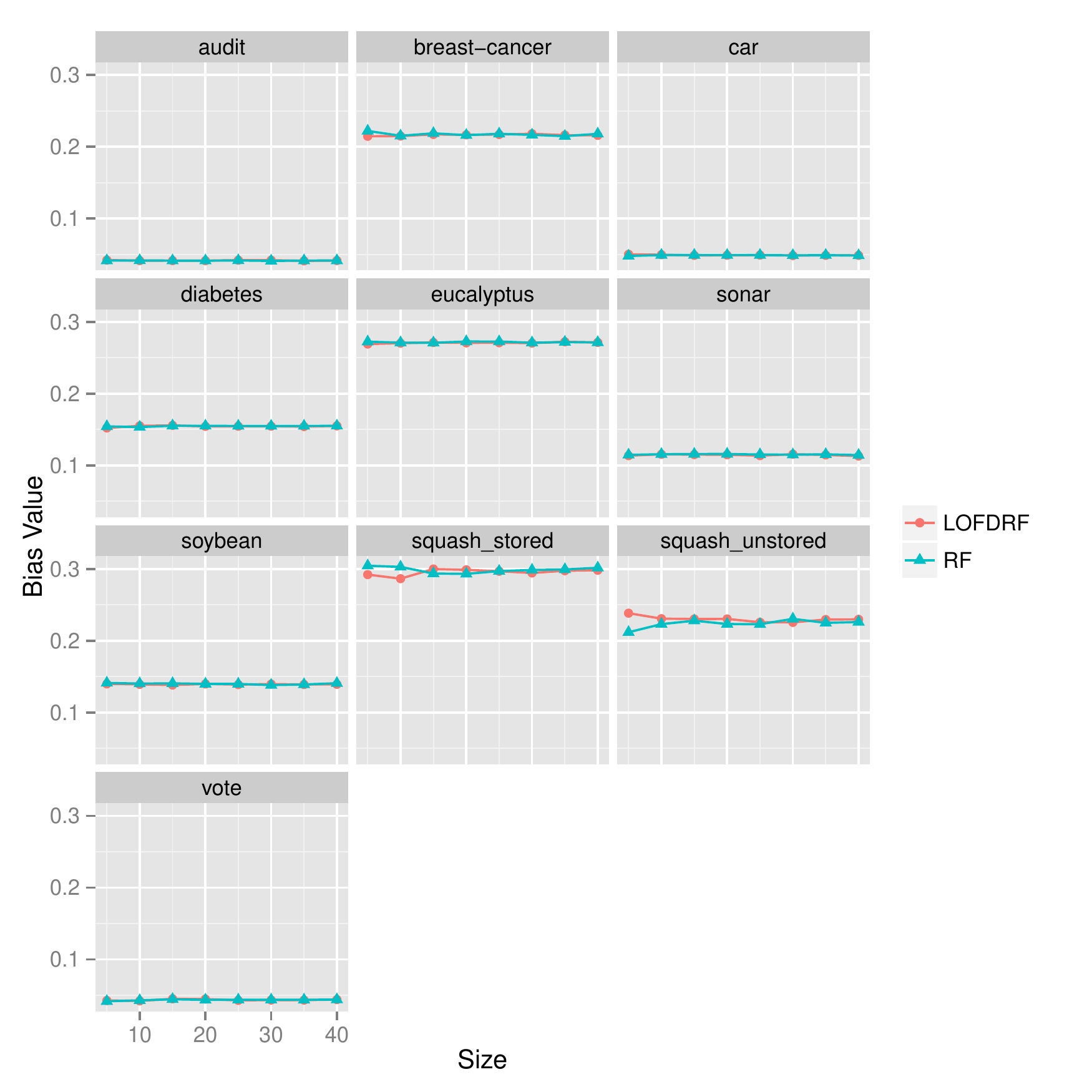}}
\caption{Bias Comparison of LOFB-DRF and Random Forest}
\label{biasComp}
\end{figure}

\begin{figure}[H]
\centering
\scalebox{1.0}{

\includegraphics[width=6in,height=4in]{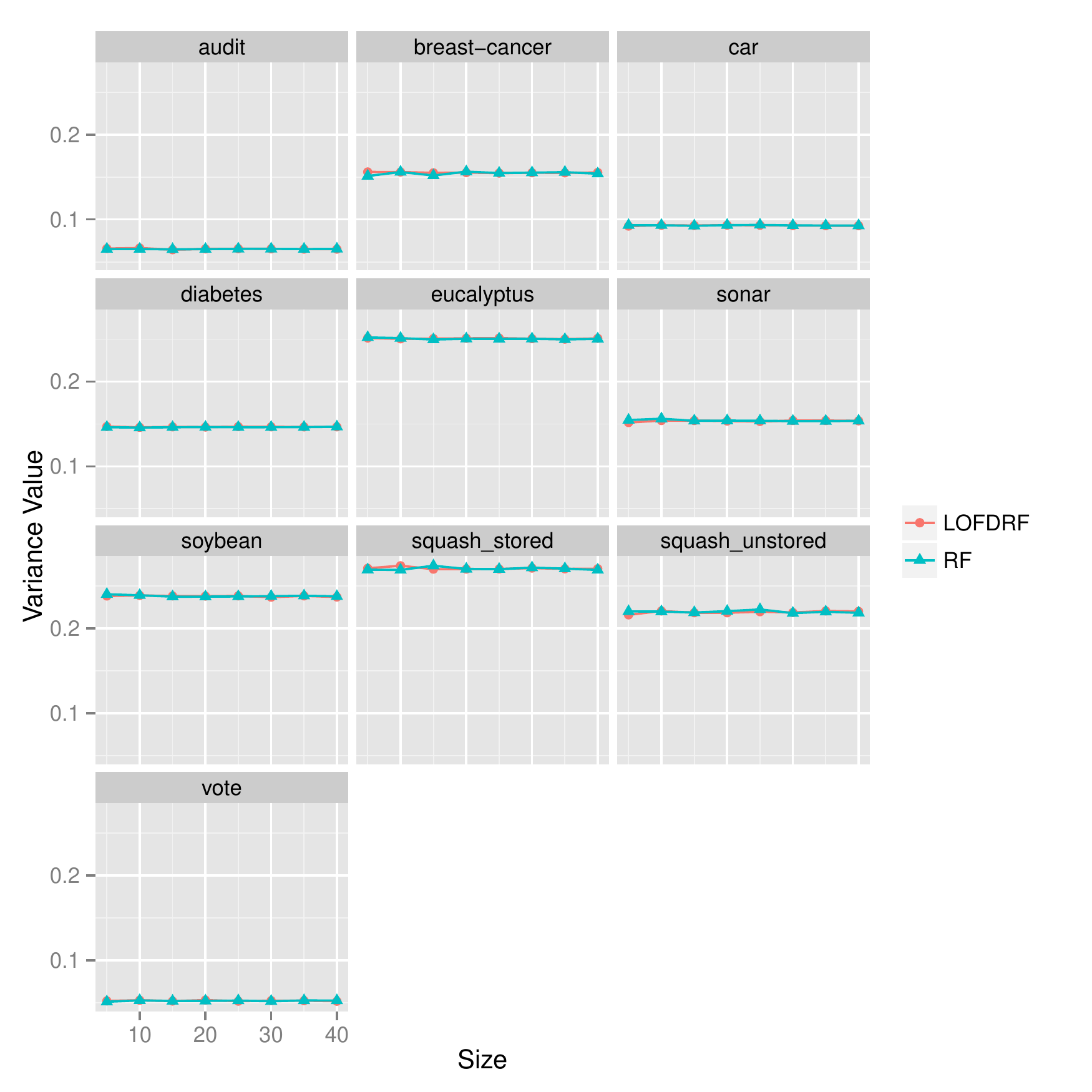}}
\caption{Variance Comparison of LOFB-DRF and Random Forest}
\label{varianceComp}
\end{figure}

We have also conducted experiments to compare the bias and variance between LOFB-DRFs and Random Forests of identical size. Figure \ref{biasComp} compares the bias and Figure  \ref{varianceComp} compares the variance. Both figures show that LOFB-DRF in most cases can have bias and variance equal to or better than Random Forest.

\section{Conclusion and Future Directions}
\label{conc}

Research conducted in this paper was based on how diversity in ensembles tends to yield better results \cite{kuncheva2003measures}  \cite{brown2005diversity} \cite{adeva2005accuracy} \cite{tang2006analysis}. We adopted the Local Outlier Factor method to select diverse trees in an RF and then used these trees to form a pruned ensemble of the original ensemble. The selection was based on both LOF value and predictive accuracy of each tree. Experimental results have shown the potential of this method with extreme pruning of Random Forests that can outperform the original population of trees with values reaching 99\% pruning level. This makes the pruned ensemble a suitable candidate for real-time applications. 

We have selected trees that correspond to the instances with the top \emph{k} weighted LOF values. Another interesting variation would be to use a hybrid approach that combines LOF with clustering to boost diversity up. Using this approach, we first create clusters of trees then from each cluster, we select a representative that corresponds to the instance with the highest weighted LOF value. The current implementation also gives equal importance to the peculiarity of the tree as measured by the LOF score and the predictive accuracy, represented by the percentage of correctly classified instances for the tree. However, tuning this significance can play an important role in enhancing the classifier. From one hand, choosing trees with higher predictive accuracy can lead to model overfitting, and on the other hand, using LOF only can lead to leaving out trees that are most representative of the dataset. Balancing between the two can result in an ensemble that is diverse enough to boost the accuracy.

%
\tiny

\begin{center}

\begin{longtable}{|l||l||l||l||l||l||l||l||l||l|}

\caption[Performance Metrics of LOFB-DRF \& RF]{Performance Metrics of LOFB-DRF \& RF}
\label{weightedlofontraintable} \\
\hline \multicolumn{1}{|c|}{\textbf{LOFB-DRF Size}} & \multicolumn{1}{c|}{\textbf{AVG}} & \multicolumn{1}{c|}{\textbf{MIN}} & \multicolumn{1}{c|}{\textbf{MAX}} & \multicolumn{1}{c|}{\textbf{SD}} & \multicolumn{1}{c|}{\textbf{Fmeasure}} & \multicolumn{1}{c|}{\textbf{AUC}} & \multicolumn{3}{|c|}{\textbf{AVG FMeasure AUC}} \\ \hline 
\endfirsthead

\multicolumn{3}{c}%
{{\bfseries \tablename\ \thetable{} -- continued from previous page}} \\
\hline \multicolumn{1}{|c|}{\textbf{LOFB-DRF Size}} & \multicolumn{1}{c|}{\textbf{AVG}} & \multicolumn{1}{c|}
{\textbf{MIN}} & \multicolumn{1}{|c|}{\textbf{MAX}} & \multicolumn{1}{|c|}{\textbf{SD}} & \multicolumn{1}{|c|}
{\textbf{Fmeasure}} & \multicolumn{1}{|c|}{\textbf{AUC}} & \multicolumn{3}{|c|}{\textbf{AVG FMeasure AUC}}
\\ \hline 
\endhead

\hline \multicolumn{3}{|r|}{{Continued on next page}} \\ \hline
\endfoot

\hline \hline
\endlastfoot
 \hline
\bf{breast-cancer} \\
 \hline
            5  &     67.01  &    61.86    &  74.23    &   3.16   &   0.65    &  0.57  &  71.13 \hspace{2.5 mm} 0.65  \hspace{3 mm} 0.58 \\
  \hline
            10   &   67.22   &   64.95    &  69.07     &  1.71    &  0.66   &   0.58 \\
 \hline
            15    &  \bf{71.34}    &  67.01    &  76.29     &  3.12   &   0.65    &  0.58 \\
 \hline
            20    &   69.48    &  67.01    &  73.20     &  2.62   &   0.66   &   0.58 \\
 \hline
            25   &    \bf{71.86}    &  69.07    &  74.23     &  1.46   &   0.65   &   0.58 \\
 \hline
            30   &    70.41   &   68.04    &  72.16    &   1.53   &   0.65   &   0.58 \\
 \hline
            35   &    70.62   &   65.98    &  73.20    &   1.91   &   0.65   &   0.58 \\
 \hline
            40   &    69.18   &   64.95    &  72.16    &   2.14    &  0.65    &  0.58 \\
 \hline
\bf{squash-unstored} \\
 \hline
     5     &   58.89   &   44.44  &    83.33  &     12.47  &    0.58  &    0.66  &  61.11  \hspace{2.5 mm}  0.52  \hspace{3 mm}   0.64 \\
\hline
            10   &    54.44   &  33.33   &   66.67   &    9.56   &   0.56   &   0.66 \\
\hline
            15   &   60.56   &   50.00   &   83.33     &  8.77   &   0.55   &   0.65 \\
\hline
            20   &    60.00   &  50.00   &   66.67    &   5.98   &   0.54   &   0.66 \\
\hline
            25   &    \bf{63.33}   &   55.56  &   77.78    &   7.93    &  0.54    &  0.65 \\
\hline
            30   &    58.33    &  44.44   &   77.78     &  8.70    &  0.53   &   0.65 \\
\hline
            35   &    \bf{67.22}    &  50.00   &   83.33    &   10.08    &  0.54    &  0.66 \\
\hline
            40   &   57.78   &   50.00   &   66.67    &   6.19    &  0.53    &  0.65 \\

 \hline

\hline
\bf{squash-stored} \\
\hline
          5   &      \bf{56.67}    &  38.89   &   66.67    &   9.56   &   0.57   &   0.59 &   55.56 \hspace{2.5 mm}  0.51  \hspace{3 mm}   0.56 \\
\hline 
           10   &    \bf{59.44}    &  44.44   &   66.67    &   7.05   &   0.54   &   0.58\\
\hline
            15    &  \bf{58.33}   &  50.00    &  66.67    &   4.48    &  0.54   &   0.58 \\
\hline
            20    &   \bf{58.33}    &  50.00    &  61.11    &   3.73    &  0.55   &   0.58 \\
\hline
            25    &  \bf{58.33}   &   50.00    &  66.67    &   5.12    &  0.53   &   0.57 \\
\hline
            30    &  \bf{56.67}   &   55.56   &   61.11    &   2.22   &   0.52   &   0.56 \\
\hline
            35    &  \bf{56.11}    &  55.56   &   61.11    &   1.67  &    0.52   &   0.57 \\
\hline
            40    &   \bf{56.11}   &   55.56  &   61.11    &   1.67   &   0.52   &   0.56 \\
 \hline
\bf{eucalyptus} \\
 \hline
  5  &      \bf{25.80}   &   11.20    &  40.40   &    8.73  &    0.26  &    0.60  &  19.92  \hspace{2.5 mm}  0.21 \hspace{3 mm}   0.57 \\
\hline
            10   &    \bf{21.00}    &  12.40   &   28.40   &    4.70   &   0.24   &   0.59 \\
\hline
            15   &    \bf{24.32}  &    14.80  &    32.00  &     5.01    &  0.24   &   0.58 \\
\hline
            20  &     \bf{24.48}   &   15.60    &  29.60  &     4.55  &    0.23   &   0.58 \\
\hline
            25    &   \bf{24.68}  &    21.20  &    29.60   &    2.35  &    0.23   &   0.58 \\
\hline
            30    &   \bf{24.80}  &    14.80   &   33.60     &  5.13    &  0.23  &    0.58 \\
\hline
            35    &   \bf{23.96}  &    20.00   &   34.40     &  4.20   &   0.23   &   0.58 \\
\hline
            40  &     \bf{21.16}  &    15.20   &   28.00   &    3.69    &  0.22  &     0.57 \\
 \hline

\bf{soybean} \\
 \hline
           5   &     77.28  &     60.78  &    85.78  &     6.80  &    0.79   &   0.88   & 77.59  \hspace{2.5 mm}  0.73 \hspace{3 mm}    0.85 \\
\hline
            10   &    \bf{78.45}  &    70.69  &    85.34    &   5.46  &    0.75  &    0.87 \\
\hline    
        15   &    \bf{79.57}   &   72.84    &  83.62     &  3.50  &     0.76   &   0.87 \\
\hline
            20 &     76.85  &    74.57   &   78.88  &     1.26 &     0.74  &    0.86 \\
\hline
            25 &     76.90  &    74.14 &     79.31   &    1.88  &    0.74   &   0.86 \\
\hline
            30   &   76.85   &   72.41  &    81.47    &   2.43    &  0.74  &    0.86 \\
\hline
            35   &   77.33  &    71.98   &   82.33   &    3.66   &   0.73  &    0.86 \\
\hline
            40  &    76.59  &    71.98   &   81.03    &   2.59   &   0.73  &    0.85 \\
 \hline

\bf{diabetes} \\
 \hline
 5   &    80.80 &     74.71  &    84.29  &     3.53  &    0.72  &    0.68 &   81.26  \hspace{2.5 mm}  0.71   \hspace{3 mm}  0.67 \\
\hline          
  10   &   81.15   &   74.71   &   84.29    &   3.56    &  0.71 &     0.68 \\
\hline
            15  &    79.85  &    77.39  &    83.14   &    1.96  &    0.71  &    0.67 \\
\hline
            20  &     \bf{81.42}   &   79.31  &    83.14  &     1.24  &    0.71 &     0.67 \\
\hline
            25  &    80.96   &   78.93    &  82.76   &    1.31   &   0.71  &    0.67 \\
\hline
            30   &   80.88    &  78.54  &    82.76   &    1.14   &   0.71   &   0.67 \\
\hline
            35   &   79.81   &   77.39   &   81.99    &   1.40    &  0.71    &  0.67 \\
\hline
            40   &    \bf{81.38}  &    80.08   &   83.14    &   0.94   &   0.71   &   0.67 \\
 \hline

\bf{car} \\
 \hline
 5   &     \bf{64.17}  &    62.41  &    67.52   &    1.33 &     0.56   &   0.78 &   62.26  \hspace{2.5 mm}  0.56  \hspace{3 mm}   0.78 \\
\hline          
  10   &    \bf{63.01}  &     61.56    &  64.29    &   0.75   &   0.56   &   0.78 \\
\hline
            15  &     \bf{62.36}  &    60.71   &   64.29  &     1.12  &    0.56  &    0.78 \\
\hline
            20  &     \bf{62.35}    &  61.22  &    63.78   &   0.82  &    0.56  &    0.78 \\
\hline
            25    &   \bf{62.69}    &  60.88    &  63.95    &   0.85   &   0.56    &  0.78  \\
\hline
            30     & 62.18    &  61.05   &   63.10    &   0.82  &    0.56  &    0.78  \\
\hline
            35   &   61.96    &  60.88   &   63.61    &   0.72  &    0.56    &  0.78 \\
\hline
            40    &  61.99    &  61.05   &   62.59   &    0.54  &     0.55 &     0.78 \\
 \hline

\bf{sonar} \\
 \hline
 5   &     \bf{12.25} &     7.04  &    18.31  &     3.34   &   0.26    &  0.00   & 0.14  \hspace{2.5 mm}  0.29  \hspace{3 mm}   0.00 \\
\hline
            10   &    \bf{9.15}  &    0.00    &  16.90   &    5.20   &   0.28  &    0.00  \\
\hline
            15   &    \bf{6.34}   &   0.00   &   14.08    &   4.47  &    0.29  &    0.00  \\
\hline
            20   &    \bf{3.38}   &   0.00   &   8.45    &   2.76    &  0.29 &     0.00 \\
\hline
            25   &    \bf{3.10}    &  0.00    &  7.04   &    2.42  &    0.28  &    0.00 \\
\hline
            30   &    \bf{1.83}   &   0.00    &  4.23   &    1.27   &   0.28   &   0.00 \\
\hline
            35   &    \bf{3.38} &     0.00  &    4.23    &   1.29 &     0.28  &    0.00 \\
\hline
            40  &     \bf{3.38}  &    0.00   &   9.86     &  2.69   &   0.28  &    0.00 \\
 \hline
\bf{audit} \\
 \hline
 5   &     95.63 &     94.26  &    96.47  &     0.72   &   0.91    &  0.89   & 96.31  \hspace{2.5 mm}  0.90  \hspace{3 mm}   0.88 \\
\hline
            10   &    95.74  &    95.00    &  96.18   &    0.35   &   0.90  &    0.88  \\
\hline
            15   &    95.99   &   95.29   &   96.47    &   0.35  &    0.90  &    0.88  \\
\hline
            20   &    96.06   &   95.29   &   96.76    &   0.39    &  0.90 &     0.88 \\
\hline
            25   &    96.22    &  95.88    &  96.47   &    0.25  &    0.91  &    0.89 \\
\hline
            30   &    96.03   &   95.59    &  96.47   &    0.25   &   0.90   &   0.88 \\
\hline
            35   &    96.26 &     95.88  &    96.47    &   0.18 &     0.90  &    0.88 \\
\hline
            40  &     96.00  &    95.59   &   96.47     &  0.27   &   0.90  &    0.87 \\
 \hline
 \hline
\bf{vote} \\
 \hline
 5   &     96.82 &     95.27  &    97.97  &     0.80   &   0.96    &  0.98   & 97.97  \hspace{2.5 mm}  0.95  \hspace{3 mm}   0.97 \\
\hline
            10   &    97.09  &    95.27    &  97.97   &    0.86   &   0.96  &    0.97  \\
\hline
            15   &    97.57   &   96.62   &   97.97    &   0.45  &    0.95  &    0.97  \\
\hline
            20   &    97.43   &   96.62   &   97.97    &   0.51    &  0.95 &     0.97 \\
\hline
            25   &    97.57    &  96.62    &  97.97   &    0.45  &    0.95  &    0.97 \\
\hline
            30   &    97.70   &   97.30    &  97.97   &    0.33   &   0.95   &   0.97 \\
\hline
            35   &    97.64 &     96.62  &    97.97    &   0.45 &     0.95  &    0.97 \\
\hline
            40  &     97.64  &    96.62   &   97.97     &  0.45   &   0.95  &    0.97 \\
 \hline
 \hline
\end{longtable}
\end{center} 
\small
\bibliographystyle{model1-num-names}
\bibliography{prl}










\end{document}